%% file: main.tex
\documentclass[conference,a4paper]{IEEEtran}
\usepackage[compatibility=false]{caption}
\IEEEoverridecommandlockouts 
\usepackage{cite}
\usepackage{amsmath,amssymb,amsfonts}
\usepackage{algorithmic}
\usepackage{graphicx}
\usepackage{textcomp}
\usepackage{framed}
\usepackage{float} 
\usepackage{cuted}
\usepackage{xcolor}
\usepackage{placeins} 
\usepackage[utf8]{inputenc}
\usepackage{pgfplots}
\DeclareUnicodeCharacter{2212}{−}
\usepgfplotslibrary{groupplots,dateplot}
\usetikzlibrary{patterns,shapes.arrows}
\pgfplotsset{compat=newest}
\usepackage[font=small,labelfont=bf]{caption} 
\usepackage{tabularx}

\usepackage{mwe} 
\usepackage{adjustbox} 
\usepackage{booktabs} 
\usepackage{lipsum}  
\usepackage{tikz}
\usepackage{multirow}   
\usepackage{hyperref}  
\usepackage{pifont}  
\usepackage{subcaption} 

\usepackage{acro}
\def\BibTeX{{\rm B\kern-.05em{\sc i\kern-.025em b}\kern-.08em
    T\kern-.1667em\lower.7ex\hbox{E}\kern-.125emX}}

\DeclareAcronym{ai}{short=AI, long= Artificial Intelligence}
\DeclareAcronym{dsp}{short=DSP, long= Digital Signal Processing}
\DeclareAcronym{sram}{short=SRAM, long= Static Random-Access Memory} 
\DeclareAcronym{sdram}{short=SDRAM, long= Synchronous Dynamic Random-Access Memory}
\DeclareAcronym{mac}{short=MAC, long= Multiply-Accumulate} 

\newcommand{\up}{\textcolor{black!60!black}{$\uparrow$}}
\newcommand{\down}{\textcolor{black!70!black}{$\downarrow$}}

\usepackage{tikz}
\usepackage{eso-pic}

\begin{document} 

\title{PicoSAM2: Low-Latency Segmentation In-Sensor for Edge Vision Applications}

\author{
    \IEEEauthorblockN{Pietro Bonazzi*$^{\dagger}$, Nicola Farronato*$^{\dagger\ddagger}$, Stefan Zihlmann*$^{\dagger}$, Haotong Qin$^{\dagger}$, Michele Magno$^{\dagger}$}
    \\
    $^{\dagger}$ETH Z\"urich, Z\"urich, Switzerland\quad
    $^{\ddagger}$IBM Research, Z\"urich, Switzerland\\
    \thanks{*Equal contribution. Corresponding author: {pbonazzi@ethz.ch}}
}

\maketitle

\AddToShipoutPictureFG*{%
  \AtPageUpperLeft{%
    \begin{tikzpicture}[remember picture,overlay]
      \node[anchor=north] at ([yshift=-2mm]current page.north){%
        \parbox{\paperwidth}{\centering \color{gray}\bfseries
          This paper has been honored with the Outstanding Lecture Award at the\\
          IEEE Sensors Conference 2025 in Vancouver, Canada%
        }%
      };
    \end{tikzpicture}
  }
}

\input{pages/0_abstract}
\input{pages/1_introduction}
\input{pages/2_related_work}
\input{pages/3_methodology}
\input{pages/4_results}
\input{pages/5_conclusion}

\bibliographystyle{IEEEtran}
\bibliography{IEEEabrv,main}

\end{document}

%% file: pages/0_abstract.tex
\begin{abstract} 
Real-time, on-device segmentation is critical for latency-sensitive and privacy-aware applications like smart glasses and IoT devices. We introduce PicoSAM2, a lightweight (1.3M parameters, 336M MACs) promptable visual segmentation model optimized for edge and in-sensor execution, including the Sony IMX500. It builds on a depthwise separable U-Net, with knowledge distillation and fixed-point prompt encoding to learn from the Segment Anything Model 2 (SAM2). On COCO and LVIS, it achieves 51.9\% and 44.9\% mIoU, respectively. The quantized model (1.22MB) runs at 14.3ms on the IMX500— achieving $\sim$86 MACs/cycle making it the only model meeting both memory and compute constraints for in-sensor deployment. Distillation boosts LVIS performance by +3.5\% mIoU and +5.1\% mAP. These results demonstrate that efficient, promptable segmentation is feasible directly on-camera, enabling privacy-preserving vision without cloud or host processing.

\end{abstract}




\section*{Funding \& Acknowledgments}

This research was funded by Swiss National Science Foundation (Grant 219943), and the European Union’s HORIZON-MSCA-2022-DN-01 (Grant 101119554).

%% file: pages/1_introduction.tex
\section{Introduction}

Recent advances in task-agnostic segmentation, such as Meta’s Segment Anything Model (SAM)\cite{kirillov2023segment} and its successor SAM2\cite{ravi2024sam2}, enable high-quality prompt-based segmentation in various visual tasks. While SAM2 \cite{ravi2024sam2} improves accuracy,  efficiency, and supports video input, its large size and transformer-based architecture make it unsuitable for deployment in latency and power constrained settings like smart cameras, wearable glasses, battery-operated device for the Internet of Things and drones.

Edge computing offers a solution by enabling low-latency, private inference directly on-device \cite{lin2020mcunet, wang2020fann, giordano2022survey, moosmann2023ultraefficient}. Recent advances in smart sensors are driving a shift toward in-sensor intelligence, enabling real-time perception directly at the point of capture.  A novel sensor realized by Sony, the IMX500 camera sensor \cite{imx500_sony_sensor, eki2021sonyIMX500, bonazzi2023tinytracker} integrates a camera with an edge AI processor \cite{zhou2020near}, enabling real-time AI at the sensor level.  However, strict hardware constraints, including a total model size below 8\,MB and limited ONNX operator support, render existing promptable segmentation models, including TinySAM \cite{shu2023tinysam}, EdgeSAM \cite{zhou2023edgesam}, MobileSAMv2 \cite{zhang2023mobilesamv2}, and LiteSAM \cite{fu2024litesam} undeployable directly on the IMX500. In fact, these models either exceed memory limits, rely on unsupported operators, or require architectural components (e.g., transformers) incompatible with the IMX500's execution engine.

This paper addresses the limitations mentioned above by introducing PicoSAM2 (see Figure~\ref{fig:picosam_architecture}), the first segmentation model to demonstrate practical in-sensor deployment designed for resource-constrained edge platforms. 

\begin{figure}[t]
    \centering
    \includegraphics[width=0.95\linewidth]{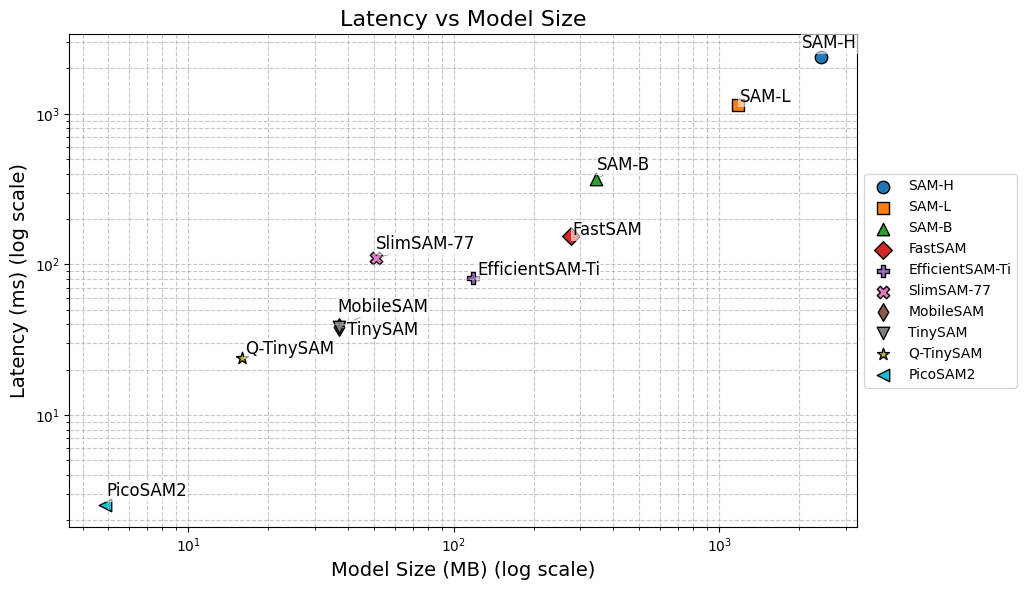}
    \caption{Comparison of segmentation models: Latency vs. Memory}
    \label{fig:latency}
\end{figure}

\begin{figure*}[!t]
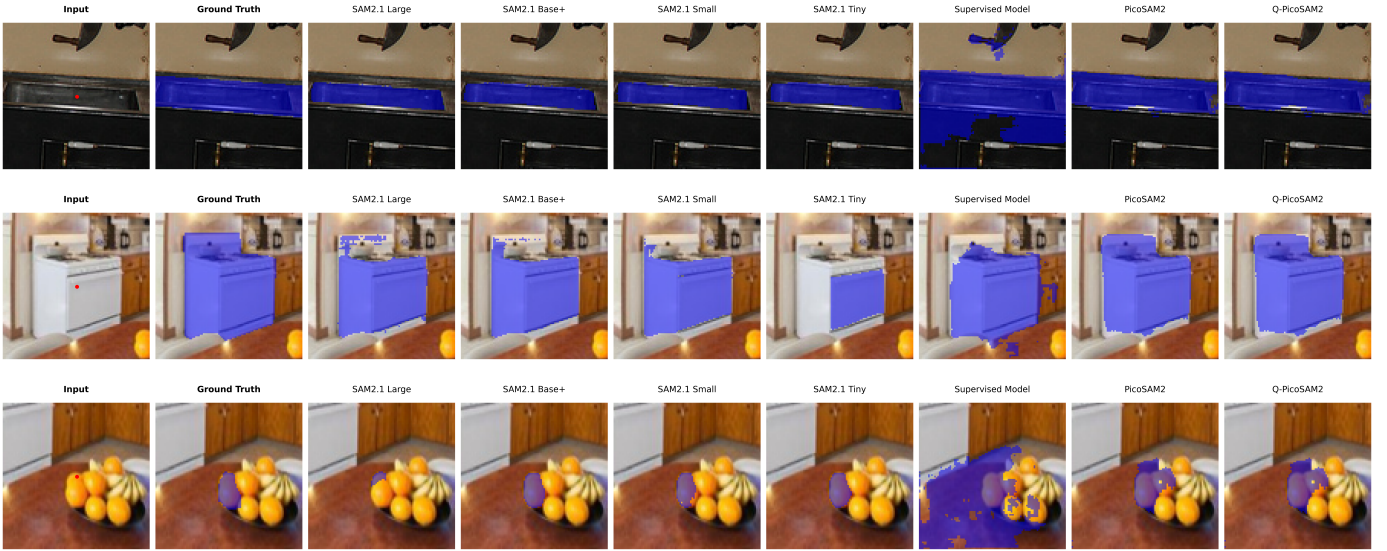

    \centering
    \includegraphics[width=\textwidth]{figures/mask_comparison_all_models18.png}\\
    \includegraphics[width=\textwidth]{figures/mask_comparison_all_models25.png}\\
    \includegraphics[width=\textwidth]{figures/mask_comparison_all_models30.png}\\[0.5em]
    \caption{Qualitative comparison of each model's mask inference.}
    \label{fig:qualitative_banner}
\end{figure*}

%% file: pages/2_related_work.tex
\section{Related Work}

Promptable visual segmentation has recently seen rapid advancement, with efforts focused on efficient training \cite{guo2023fastsam, fu2024litesam}, knowledge distillation \cite{chen2023mobilesam, zhou2023edgesam, li2024efficientsam, zhang2024efficientvit, liu2024repvit, wang2024samlightening, shu2023tinysam}, model pruning \cite{kim2024slimsam}, and training-free methods \cite{park2024expeditsam} to enable efficient architectures \cite{zhao2023fast} that make powerful models like SAM \cite{kirillov2023segment} deployable on edge devices. TinySAM \cite{shu2023tinysam} distills SAM into a smaller and faster model via stage-wise supervision and prompt sampling, followed by post-training quantization \cite{banner2019post, nagel2020up}. Our work builds on this by adopting a fully custom U-Net-style student without relying on the original TinySAM backbone. EdgeSAM \cite{zhou2023edgesam} replaces the SAM transformer encoder with a CNN, achieving over 30 FPS on mobile devices. Their “prompt-in-the-loop” distillation iteratively refines student masks with feedback prompts. In contrast, our approach encodes fixed prompt behavior statically, eliminating run-time prompt overhead. SAM2 \cite{ravi2024sam2} generalizes prompt segmentation to video using a pyramid transformer backbone \cite{vasudevan2023hiera}. We draw inspiration from its architecture while focusing exclusively on static images and remove temporal components to enable lightweight deployment. MobileSAMv2 \cite{zhang2023mobilesamv2} optimizes the SAM decoder by reducing the prompt complexity and simplifying the architecture. Similarly, we redesign the decoder using CNN-based mask heads and avoid expensive attention-based modules. LiteSAM \cite{fu2024litesam} demonstrates high performance with lightweight modular components, including a transformer-based LiteViT encoder \cite{wang2023litevit} and AutoPPN \cite{xu2023auto}. Although effective, its reliance on transformer blocks limits its suitability for MCU deployment. Instead, we propose a dense CNN backbone with pyramid features to mimic transformer-like capacity in-sensor \cite{imx500_sony_sensor, eki2021sonyIMX500}.

%% file: pages/3_methodology.tex
\section{Methodology}

Adapting the high-performance SAM2 model for edge and in-sensor inference—such as on the Sony IMX500 \cite{imx500_sony_sensor}—presents three key challenges: a strict 8MB memory limit, restricted ONNX operator support, and RGB-only input that excludes multimodal prompting. This paper proposes and evaluates a novel segmentation design to address these constraints.

\subsection{Model Architecture}

We designed a lightweight U-Net \cite{ronneberger2015unet} with separable convolutions in depth \cite{chollet2017xception}, skip connections, and up/down sampling ensuring spatial preservation and quantization-friendliness. 

\begin{figure}[htbp]
    \centering
    \includegraphics[width=0.95\linewidth]{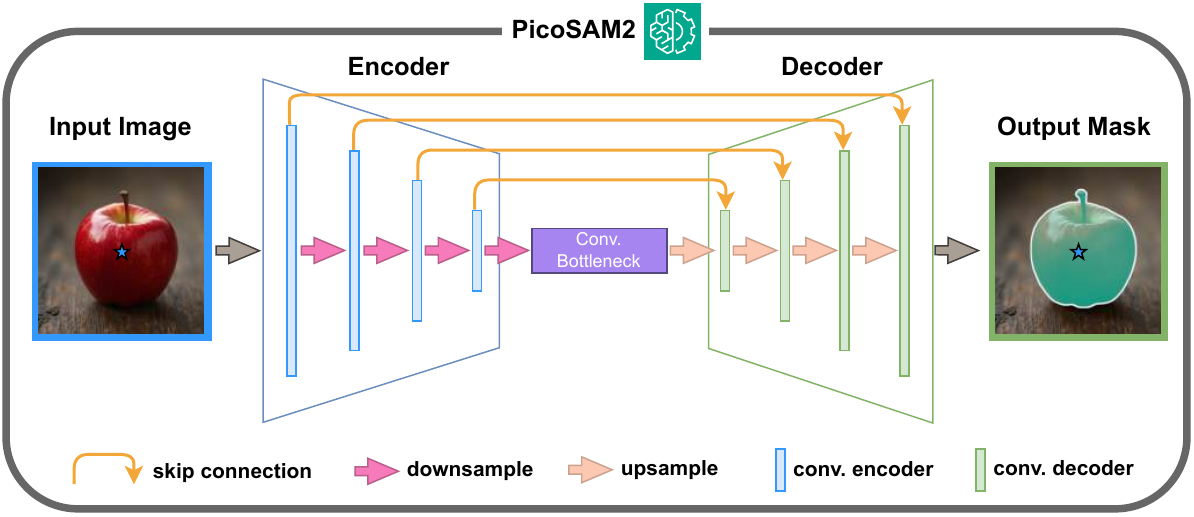}
    \caption{Schematic of the PicoSAM2 architecture.}
    \label{fig:picosam_architecture}
\end{figure}

Since we restrict inputs to RGB channels only, explicit prompt encoding is not feasible. To overcome this, we crop each training image such that the prompt point is centered, enabling the network to learn a spatial prior. This implicit mechanism allows the model to be promptable without additional input channels: the student observes only RGB crops while the teacher provides supervision with full prompts.

\subsection{Knowledge Distillation}

Our training objective combines soft supervision from the teacher model with hard supervision from ground truth masks. Specifically, we use a two-part loss: (1) a mean squared error loss ($\mathcal{L}_{\text{MSE}}$) to match the teacher’s soft logits \cite{hinton2015distilling}, and (2) a dual-component alignment loss consisting of balanced binary cross-entropy ($\mathcal{L}_{\text{BCE}}$) and Dice loss ($\mathcal{L}_{\text{Dice}}$) \cite{ronneberger2015unet, milletari2016vnet}. The total loss is:
\begin{align}
\mathcal{L}_{\text{total}} &= \lambda \cdot \mathcal{L}_{\text{MSE}} + (1 - \lambda) \cdot \left(0.5\,\mathcal{L}_{\text{BCE}} + 0.5\,\mathcal{L}_{\text{Dice}}\right)
\end{align}
where $\lambda \in (0, 1)$ is dynamically adjusted based on the confidence of the teacher \cite{Sanh2019DistilBERTAD}, allowing the student to rely more heavily on the teacher when the predictions are reliable and on the ground truth otherwise.

\subsection{Training, Quantization and Deployment}

The student model was trained on COCO \cite{lin2014microsoft} using the AdamW optimizer \cite{loshchilov2017decoupled} and evaluated on LVIS \cite{gupta2019lvis}. After training, we perform static quantization to INT8 \cite{gholami2021surveyquantizationmethodsefficient, habi2021hptq, gordon2024eptq, dikstein2025dgh, torchvision_models, keras_applications}. For hardware benchmarking, we used the Sony IMX500 intelligent vision sensor \cite{eki2021sonyIMX500, imx500_sony_sensor}, a stacked BSI-CMOS image sensor with an integrated DSP optimized for CNN inference. The sensor features 2304 MAC units and a 262.5MHz programmable DSP capable of 4.97 TOPS/W. CNN results were transmitted via the SPI interface, enabling a low-power, compact edge deployment, demonstrating that promptable segmentation can run in-sensor under tight memory and operator constraints.

%% file: pages/4_results.tex
\begin{table*}[!t]
\centering
\caption{Comparison of Edge Segmentation Models. Latencies were measured on a NVIDIA T4 instance (GPU Latency) and Sony IMX500 (DPU Latency) where available.}
\label{tab:merged_results}
\begin{tabularx}{\textwidth}{X
                                r
                                c
                                c
                                r
                                r
                                r
                                r}
\toprule
\textbf{Model} & \textbf{MACs \down} & \textbf{COCO mIoU \up} & \textbf{LVIS mIoU \up} & \textbf{Params (M) \down} & \textbf{Size (MB) \down} & \textbf{GPU Latency \down} & \textbf{DPU Latency \down} \\
\midrule
SAM-H             & 2.97T  & 53.6 & 60.5 & 635   & 2420.3 & 2.39\,s     & -- \\
FastSAM           & 443G   & 51.6 & 55.2 & 72.2  & 275.6  & 153.6\,ms   & -- \\
TinySAM           & 42G    & 50.9 & 52.1 & 9.7   & 37.0   & 38.4\,ms    & -- \\
EdgeSAM           & 42G    & 48.0 & 53.7 & 9.7   & 37.0   & 38.4\,ms    & -- \\
LiteSAM           & 4.2G   & 49.0 & 51.2 & 4.2   & 16.0   & 16.0\,ms    & -- \\
Supervised        & 336M   & 53.0 & 41.4 & 1.3   & 4.87   & 2.5\,ms     & -- \\
\textbf{PicoSAM2} & \textbf{336M} & \textbf{51.9} & \textbf{44.9} & \textbf{1.3} & \textbf{4.87} & \textbf{2.54\,ms} & -- \\
\textbf{Q-PicoSAM2} & \textbf{324M} & \textbf{50.5} & \textbf{45.1} & \textbf{1.3} & \textbf{1.22} & \textbf{--} & \textbf{14.3\,ms} \\
\bottomrule
\end{tabularx}
\end{table*}

\begin{figure*}[!htbp]
\centering
\begin{subfigure}{0.49\linewidth}
    \includegraphics[width=\linewidth]{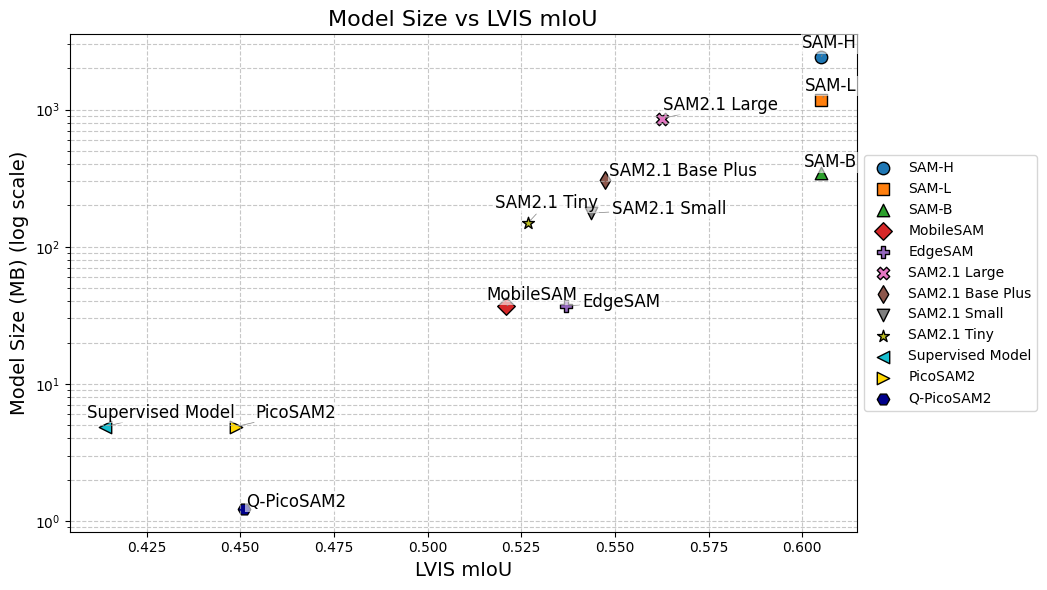}
    \caption{LVIS mIoU vs. Size}
    \label{fig:lvis_miou_vs_size}
\end{subfigure}
\hfill
\begin{subfigure}{0.49\linewidth}
    \includegraphics[width=\linewidth]{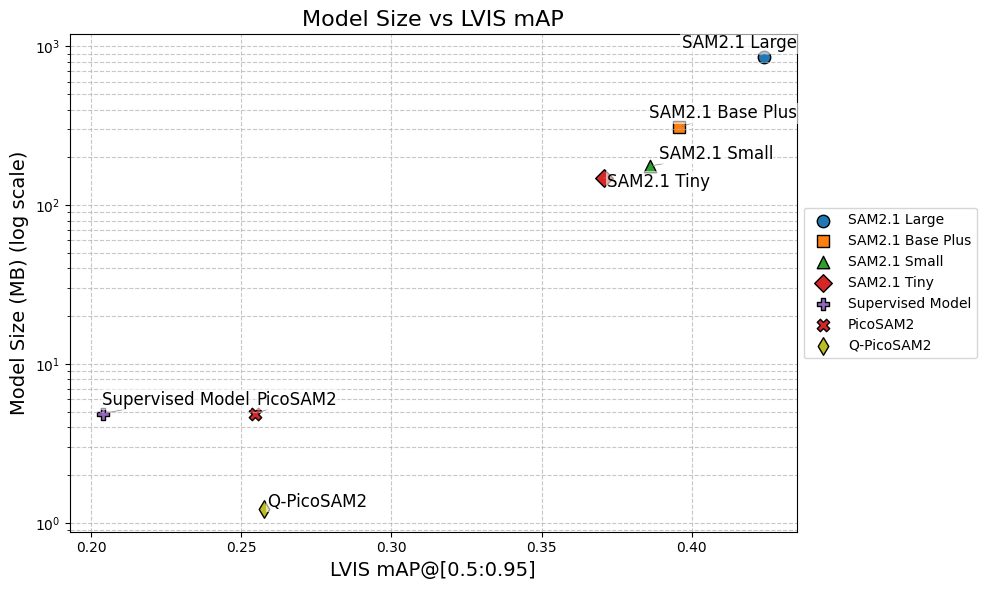}
    \caption{LVIS mAP vs. Size}
    \label{fig:lvis_map_vs_size}
\end{subfigure}
\caption{Segmentation accuracy (mIoU) and precision (mAP) vs. model size on LVIS (log scale).}
\label{fig:lvis_results}
\end{figure*}

\section{Experimental Results}

PicoSAM2 was benchmarked against lightweight segmentation models, see Table~\ref{tab:merged_results}, evaluating compute cost, accuracy, and deployability. For comparison, all the results, except PicoSAM2, were taken from the TinySAM paper~\cite{shu2023tinysam}.
Accuracy of PicoSAM2 was measured using mIoU with a single centered point prompt on cropped images, ensuring the prompt target is the central object.

PicoSAM2 achieves the lowest compute cost at 336M MACs and an inference latency of 2.54ms on an NVIDIA T4 GPU. Q-PicoSAM2 runs at 14.3ms inference latency on the Sony IMX500. Based on this, it achieves approximately 86 MACs per cycle on the IMX500. It is also the only segmentation model to satisfy the strict $<8$\,MB memory constraint (1.3M parameters, 1.22MB quantized). Despite these constraints, it delivers 51.9\% mIoU on COCO and 44.9\% on LVIS.

Distillation yields a boost of +3. 5\% mIoU and +5. 1\% mAP in LVIS versus standard training. Figure~\ref{fig:latency} illustrates the trade-off between latency and model size on a logarithmic scale, highlighting the position of PicoSAM2 at the lowest latency and smallest size among all models compared. PicoSAM2 achieves strong segmentation accuracy (mIoU) on COCO and LVIS datasets relative to its size (see Figure~\ref{fig:lvis_miou_vs_size}). The distilled PicoSAM2 model significantly improves precision in LVIS compared to the supervised baseline (see Figure~ \ref{fig:lvis_map_vs_size}). 

The qualitative comparisons in Figure~\ref{fig:qualitative_banner} show that PicoSAM2 produces high-quality single-prompt segmentation masks, thanks to its task-specific design.



%% file: pages/5_conclusion.tex
\section{Conclusion \& Discussion}

This work introduces PicoSAM2, a prompt-based segmentation model tailored for real-time deployment on the Sony IMX500 edge AI vision sensor. By rethinking the design of segmentation architectures for extreme efficiency, PicoSAM2 achieves competitive performance in three core dimensions: computational cost, segmentation accuracy, and deployability. In terms of compute, PicoSAM2 operates with just 336M MACs—less than 0.02\% of SAM-H—and achieves real-time performance with 2.5\,ms GPU latency and 14.3\,ms on-device latency after INT8 quantization. Despite its small footprint (1.3M parameters and 1.22\,MB quantized), the model retains strong performance with 51.9\% mIoU on COCO and 45.1\% on LVIS using a single-point prompt. Crucially, it meets the IMX500’s stringent 8\,MB memory restriction, operates with input only in RGB, and complies with limited operator support. Q-PicoSAM2 demonstrates high hardware efficiency, reaching $\sim$86 MACs per cycle on the Sony IMX500. Ablation studies confirm that distillation from SAM2 consistently boosts generalization over fully supervised training (+5.08\% mAP), underscoring the effectiveness of hybrid supervision in low-data regimes. Despite limited flexibility, fixed spatial prompting ensures robust promptability under severe modality constraints. This work demonstrates the viability of deploying intelligent, prompt-based vision models directly on constrained sensors.